\documentclass{article}

\usepackage{PRIMEarxiv}

\usepackage[utf8]{inputenc} 
\usepackage[T1]{fontenc}    
\usepackage{hyperref}       
\usepackage{url}            
\usepackage{booktabs}       
\usepackage{amsfonts}       
\usepackage{nicefrac}       
\usepackage{microtype}      
\usepackage{lipsum}
\usepackage{fancyhdr}       
\usepackage{graphicx}       
\graphicspath{{media/}}     

\title{
Beyond Morphology: Quantifying the Diagnostic Power of Color Features in Cancer Classification
}

\author{
  Farnaz Kheiri \\
  Dept. of Electrical, Computer and Software Engineering \\
  Ontario Tech University \\
  Oshawa, Ontario, Canada\\
  \texttt{farnaz.kheiri@ontariotechu.net} \\
   \And
  Shahryar Rahnamayan \\
  Dept. of Engineering  \\
  Brock University \\
  St. Catharines, Ontario, Canada\\
  \texttt{srahnamayan@brocku.ca} \\
  \And
  Masoud Makrehchi \\
  Dept. of Electrical, Computer and Software Engineering \\
  Ontario Tech University \\
  Oshawa, Ontario, Canada\\
  \texttt{Masoud.Makrehchi@ontariotechu.ca} \\
}

\begin{document}
\maketitle

\begin{abstract}
In histopathology, human experts primarily rely on color as a means of enhancing contrast to interpret tissue morphology, whereas machine vision models process color as raw statistical information. This distinction raises a fundamental question: to what extent can pixel intensity alone, independent of structural and morphological cues, support cancer classification? To address this question, we systematically evaluated the standalone discriminative power of global color features while deliberately excluding all morphological information. Specifically, we extracted statistical color moments and discretized RGB and HSV color histograms, and assessed their performance across ten diverse experimental settings using classical machine learning classifiers. Our results demonstrate that color features alone can achieve strong performance in binary diagnostic tasks (e.g., benign versus malignant), with classification accuracies reaching up to 89\%. This performance is likely attributable to global chromatic shifts associated with malignancy. Importantly, these simple color-based representations consistently outperformed random baselines by a substantial margin, indicating that raw color distributions encode a non-random and diagnostically relevant signal for cancer detection. Consequently, this study suggests that simple, computationally efficient color features can serve as an effective pre-screening tool. By identifying samples with strong chromatic indicators of malignancy, these lightweight models could function as a first-pass triage system, reducing the computational burden on complex deep learning architectures.
\end{abstract}


\section{Background and Literature Review}
Histopathological data typically refer to microscopic images of human tissues obtained through biopsy or surgical resection. Before these images can be analyzed, the tissue undergoes a staining procedure \cite{mezei2024image}. Staining is a laboratory technique that adds color to tissue sections that are naturally transparent allowing different cellular structures to become visible under a microscope. Because unstained tissue is visually homogeneous, pathologists and computational models cannot easily distinguish nuclei, cytoplasm, connective tissue, or pathological abnormalities such as cancer. During staining, different cellular components absorb dyes at varying intensities, and these chemical interactions enhance the contrast between biological structures, making diagnostic features more discernible \cite{golberg2024application, javaeed2021histological}. Among the various staining techniques used in diagnostic pathology, Hematoxylin and Eosin (H\&E) is the most widely adopted. H\&E provides a clear and consistent contrast between nuclei and cytoplasmic regions, hematoxylin stains nuclei blue-purple, while eosin stains cytoplasm and extracellular matrix pink. So, this makes it the standard stain for cancer diagnosis and the primary format in most histopathology imaging datasets \cite{dunn2024quantitative}.

These stained images represent the foundation for diagnosing many diseases, particularly cancer, and are routinely interpreted by both human pathologists and machine learning models. To support research in computational pathology, several large-scale histopathology datasets have been made publicly available. Prominent examples include TCGA \cite{tomczak2015review}, Camelyon16/17 \cite{bejnordi2017diagnostic, bandi2018detection}, PCam \cite{Veeling2018-qh}, and BreakHis \cite{spanhol2015dataset}, all of which provide gigapixel whole-slide images (WSIs) captured at high magnification. Because WSIs are extremely large, they are commonly partitioned into smaller patches (e.g., 224×224 or 512×512 pixels) to facilitate computational analysis and model training.

Pathologists assess stained histopathology slides by examining the tissue’s microscopic architecture and cellular morphology to identify abnormalities indicative of cancer. Using stains such as H\&E, they evaluate key features including nuclear size and shape, chromatin texture, mitotic activity, glandular or tissue organization, and the presence of necrosis or invasion into surrounding structures. These visual cues help distinguish normal from malignant tissue and differentiate between cancer types and subtypes, such as adenocarcinoma, squamous cell carcinoma, or lymphoma. Pathologists integrate these microscopic observations with clinical information to determine tumor grade, aggressiveness, and diagnostic classification \cite{jahn2020digital, madabhushi2016image}.

In contrast, machine vision models learn cancerous features through data-driven pattern recognition rather than explicit morphological reasoning. Deep learning systems, such as convolutional neural networks and transformer-based architectures, analyze thousands of image patches to automatically extract hierarchical representations of texture, color, nuclear arrangement, and tissue architecture without being explicitly instructed about pathology concepts \cite{kheiri2025investigation}. While pathologists deliberately focus on biologically meaningful cues, such as nuclear atypia, glandular distortion, and invasive boundaries, machine learning models may rely on both relevant cancer morphology and unintended correlations within the data \cite{coudray2018classification}. 

One of the most powerful, yet frequently overlooked, of these correlations is color. While pathologists view color merely as a tool to create contrast, a way to light up the structures they actually care about, computer models see color very differently. To a machine, color is raw numerical data. And in the context of cancer detection, this raw color signal can be surprisingly informative, for different reasons. First, there is a biological reason. Cancerous cells are often more absorbent of dye than healthy cells. Malignant nuclei tend to contain more DNA, which absorbs the hematoxylin stain more aggressively. This means that, statistically, a patch of cancerous tissue might simply look bluer or darker than a patch of healthy tissue. Theoretically, a computer could detect this shift in color distribution without knowing anything about the shape of the cell \cite{gurcan2009histopathological, dunn2024quantitative}. Second, there is a technical reason, which is far more problematic. Histopathological data sets are often merged from different hospitals. Each lab has its own recipe for staining. Some use older chemicals, some dip the slides for longer, and some digital scanners are calibrated differently. If a specific cancer type in a dataset happens to come primarily from one hospital with a distinct staining style, the AI might learn to recognize the lab's signature rather than the disease. It learns a "shortcut" based on color histograms rather than tissue architecture \cite{tellez2019quantifying, kheiri2024feature}.

Despite the explosion of research into complex Deep Learning architectures, few studies have isolated the impact of simple color features on model performance. It remains unclear whether high classification accuracy requires complex analysis of tissue shape, or if it can be achieved solely through the analysis of pixel intensity. To address this, this paper evaluates the standalone predictive power of color by intentionally excluding all morphological and textural information.

Our experimental results demonstrate that high classification accuracy is indeed achievable using only global color features, with performance reaching as high as 89\% in binary diagnostic tasks and 74\% in multi-class tissue classification. By proving that accurate classification does not always require heavy training procedures or complex feature extraction, we show that the computational burden of the diagnostic pipeline can be significantly reduced. This eliminates the dependency on high-performance GPUs which enables the deployment of lightweight algorithms on resource-constrained hardware, a shift that aligns with the principles of TinyML, allowing diagnostic models to operate directly on embedded systems with minimal latency and power consumption.

On the other hand, these findings challenge the common assumption that high diagnostic performance in histopathology must necessarily rely on complex morphological representations. Our results underscore color as a fundamental and interpretable source of diagnostic information that is often implicitly exploited by modern learning-based models, sometimes without explicit acknowledgment. By quantifying the contribution of color alone, this work provides valuable insights into the role of staining-related signals and highlights the need for greater awareness of color-driven effects in the design, evaluation, and interpretation of computational pathology systems.
The role of color in computational pathology has been examined through two distinct methodological perspectives. The first and most predominant body of literature considers color primarily as a source of domain shift and variability. They views color as an artifact of staining inconsistencies that must be corrected or normalized to allow models to focus on tissue morphology. The second, smaller body of work views color as a diagnostic signal, exploring how color features can be explicitly engineered to improve classification performance.
\begin{figure*}[ht]
    \centering
    \includegraphics[width=\textwidth]{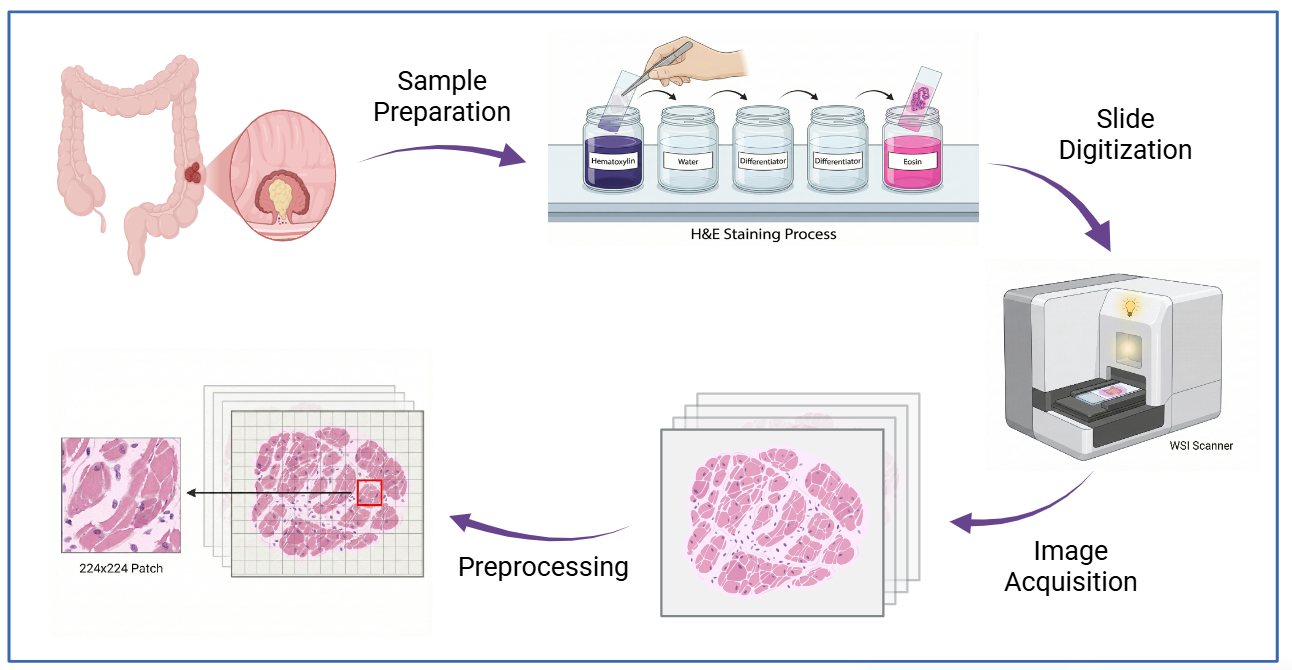} 
    \caption{The end-to-end workflow for histopathological image acquisition and data preparation. The process begins with Sample Preparation, where tissue is harvested (e.g., from the colon) and mounted on glass slides. The slides undergo the H\&E Staining Process, using Hematoxylin and Eosin to introduce chromatic contrast between nuclei and cytoplasm. These stained slides are then subjected to Slide Digitization using a high-throughput WSI scanner. Following Image Acquisition, the resulting gigapixel whole slide images undergo Preprocessing, where they are tiled into non-overlapping grids to facilitate the extraction of 224$\times$224 patches for computational analysis.}
    \label{fig:staining_workflow}
\end{figure*}
\subsection{Color as a Source of Variability (Stain Normalization)} A significant portion of histopathology research focuses on mitigating the "batch effects" caused by different staining protocols. Since deep learning models are notably sensitive to domain shifts, standard pipelines often employ stain normalization techniques to standardize the appearance of slides before training \cite{howard2021impact}. Classic algorithms, such as those proposed by Reinhard et al \cite{reinhard2002color}, Macenko et al. \cite{macenko2009method} and Vahadane et al. \cite{vahadane2016structure}, mathematically project images into a standard color space to force a uniform appearance across datasets. Unlike the classical methods which simply shift pixel values based on a formula, most recent approaches Generative Adversarial Networks (GANs) use deep learning to repaint the image. They learn the specific style of a target domain and apply it to a new image while trying to preserve the original tissue structure \cite{zanjani2018stain,shaban2019staingan, kim2024stain,cong2021semi}. Parallel to GANs, Convolutional Autoencoders (CAEs) have also been utilized for this task. CAEs encode the tissue structure into a compressed latent space and stripping away color information and then reconstructing the image using the specific color statistics of a target template \cite{janowczyk2017stain,zanjani2018stain}. 
Similarly, color augmenttion introduces extreme color variability during the training phase rather than standardizing the input data to a fixed reference. This technique, often referred to as domain randomization involves applying stochastic transformations to the brightness, contrast, hue, and saturation of image patches \cite{tellez2019quantifying,marini2023data,manuel2022impact}.
\subsection{Color as a Diagnostic Feature} Conversely, a subset of literature argues that color information, when appropriately processed, contains high-level diagnostic value. Unlike normalization approaches that seek to suppress color variance, these studies utilize color as a primary feature for classification. Early work in computational pathology demonstrated that simple statistical moments of color channels could effectively distinguish tumor tissue from normal tissue. For instance, Tabesh et al. \cite{tabesh2007multifeature} utilized a combination of color, texture, and morphometric features to diagnose prostate cancer. Their analysis revealed that global color channel statistics (such as the mean and standard deviation of pixel intensity) were significant discriminators, as malignant regions exhibited distinct spectral properties due to increased cellular density and nuclear staining. Similarly, Aswathy and Jagannath \cite{aswathy2021svm} applied this feature fusion approach to breast cancer classification by explicitly calculating statistical color features to quantify stain absorption. By integrating these values with geometrical and textural descriptors, they demonstrated that the inclusion of color intensity data significantly improved the performance of Support Vector Machine classifiers. Expanding on this, Wahyuni et al. \cite{wahyuni2022analysis} achieved higher accuracy in detecting Invasive Ductal Carcinoma by stacking histogram statistics from both RGB and HSV domains alongside GLCM texture features. In another study \cite{atrey2023multi}, Genetic Algorithms is applied to optimize a high-dimensional feature pool in breast cancer classification, which results in achieving high accuracy by explicitly retaining color descriptors alongside texture. Furthermore, recent hybrid frameworks explicitly validate the diagnostic power of low-level chromatic features. Studies \cite{joseph2022improved,hassan2022hybrid} demonstrate that integrating Colored Histograms with texture descriptors or deep learning features consistently yields high classification accuracy.

\section{Method}
In this section, we present the experimental framework designed to evaluate the standalone discriminatory power of color in histopathological analysis. Here, our methodology isolates chromatic information to quantify its specific contribution to classification performance. The proposed workflow consists of three primary stages: (1) the acquisition of diverse histopathology datasets, (2) the extraction of statistical color descriptors across multiple color spaces (RGB, HSV), and (3) the benchmarking of these features using standard machine learning classifiers (SVM, KNN, Random Forest). By intentionally excluding structural features, these experiments allow for an analysis of the predictive power of color features in cancer diagnosis.
\subsection{Dataset Description}
To ensure the generalizability of our findings across different tissue types and staining protocols, we utilized four distinct histopathology databases: PathMNIST (colorectal) \cite{yang2023medmnist}, BreakHis (breast) \cite{spanhol2015dataset}, LungHist700 (lung) \cite{diosdado2024lunghist700}, and IDC (breast) \cite{cruz2014automatic}. These datasets were further partitioned into specific experimental configurations to test classification performance across binary, multi-class, and magnification-dependent tasks.
\subsubsection{PathMNIST} The PathMNIST dataset consists of H\&E-stained histological images of colorectal cancer. We utilized this data for two distinct experimental frameworks. First, we evaluated multi-class performance using the original nine tissue phenotypes: Adipose, Background, Debris, Lymphocytes, Mucus, Smooth Muscle, Normal Colon Mucosa, Cancer-Associated Stroma, and CRC Epithelium. Second, to assess binary diagnostic capability, we aggregated these labels into two super-classes, mapping Adipose, Background, Mucus, Smooth Muscle, and Normal Colon Mucosa to the 'Normal' category, while grouping Debris, Lymphocytes, Cancer-Associated Stroma, and CRC Epithelium under the 'Abnormal' designation \cite{yang2023medmnist}. 
\subsubsection{BreakHis} We employed the Breast Cancer Histopathological Image Classification (BreakHis) dataset to analyze classification performance across three magnification levels (40×, 100×, and 200×). For each subset, we defined two distinct experimental tasks: a binary classification distinguishing between Benign and Malignant tumors, and a multi-class challenge categorizing malignant samples into four specific subtypes (Ductal, Lobular, Mucinous, and Papillary Carcinoma) \cite{spanhol2015dataset}.
\subsubsection{LungHist700} To evaluate performance on lung tissue, we utilized the LungHist700 dataset, which contains 700 high-resolution H\&E-stained images. This dataset was used for a three-class classification task comprising: Benign, Adenocarcinoma, and Squamous Cell Carcinoma \cite{diosdado2024lunghist700}.
\subsubsection{IDC (Invasive Ductal Carcinoma)}
The IDC dataset consists of H\&E-stained patches specifically focused on the detection of Invasive Ductal Carcinoma. This was utilized as a binary classification task to distinguish IDC positive patches from Normal tissue \cite{cruz2014automatic}.

\begin{table}[!ht]
    \centering

    \label{tab:dataset_summary}
    \resizebox{\textwidth}{!}{
    \begin{tabular}{lllll}
        \toprule
        \textbf{Dataset} & \textbf{Organ} & \textbf{Magnification} & \textbf{Task Type} & \textbf{Classes (Labels)} \\ 
        \midrule
        PathMNIST & Colon & Mixed & Multi-class & 9 Tissue Types \\ 
        PathMNIST (Binary) & Colon & Mixed & Binary & Normal vs. Abnormal \\ 
        \midrule
        BreakHis 40$\times$ & Breast & 40$\times$ & Binary & Benign vs. Malignant \\ 
        BreakHis 40$\times$ & Breast & 40$\times$ & Multi-class & 4 Malignant Subtypes \\ 
        \midrule
        BreakHis 100$\times$ & Breast & 100$\times$ & Binary & Benign vs. Malignant \\ 
        BreakHis 100$\times$ & Breast & 100$\times$ & Multi-class & 4 Malignant Subtypes \\ 
        \midrule
        BreakHis 200$\times$ & Breast & 200$\times$ & Binary & Benign vs. Malignant \\ 
        BreakHis 200$\times$ & Breast & 200$\times$ & Multi-class & 4 Malignant Subtypes \\ 
        \midrule
        LungHist700 & Lung & High Res & Multi-class & Benign, Adenocarcinoma, SCC \\ 
        \bottomrule
    \end{tabular}%
    }
        \caption{Summary of the histopathology datasets and experimental configurations used in this study, detailing the target organ, magnification levels, and specific classification tasks (Binary vs. Multi-class).}
\end{table}
\subsection{Color Feature Extraction Framework} To evaluate the diagnostic potential of color, our feature extraction framework is designed to discard all spatial and morphological information. We treat each histopathology image as an unordered collection of pixels, extracting global statistics that quantify stain intensity and chromatic distribution. This approach ensures that any classification success is driven solely by color properties, independent of tissue architecture. The framework operates across two distinct representations: the native RGB space and the HSV color space. 
\subsubsection{Color Moments} To capture the global color distribution with a compact feature vector, we utilized the method originally proposed by Stricker and Orengo \cite{stricker1995similarity}. This approach assumes that the color distribution of an image can be effectively characterized by its lower-order probability moments. Since probability distributions are uniquely defined by their moments, retaining the first three moments provides a sufficient approximation of the color content without requiring the storage of the full histogram. 

For an image with $N$ pixels, we extract three statistical descriptors for each color channel $i$ (where $i \in \{R, G, B\}$):
\begin{enumerate}
    \item Mean ($\mu_i$): Represents the average intensity of the color channel, quantifying the central tendency of the pixel distribution and serving as a global measure of brightness or concentration for that specific color component.
    \begin{equation}
    \mu_i = \frac{1}{N} \sum_{j=1}^{N} p_{ij}
    \end{equation}
    where $N$ is the total number of pixels in the image patch, and $p_{ij}$ represents the intensity value of the $j$-th pixel in color channel $i$.
    
    \item Standard Deviation ($\sigma_i$): Captures the contrast or spread of the color distribution, indicating the variability of staining within the tissue sample.
    \begin{equation}
    \sigma_i = \sqrt{\frac{1}{N} \sum_{j=1}^{N} (p_{ij} - \mu_i)^2}
    \end{equation}
    where $\mu_i$ is the mean calculated in Eq. 1, $N$ is the total number of pixels, and $p_{ij}$ is the intensity of the $j$-th pixel in channel $i$.

    \item Skewness ($s_i$): Measures the asymmetry of the color values, indicating whether the pixel distribution leans more toward the bright or dark regions of the intensity range. We calculate this using the cube root to ensure the result remains on the same scale as the mean and standard deviation.
    \begin{equation}
    s_i = \sqrt[3]{\frac{1}{N} \sum_{j=1}^{N} (p_{ij} - \mu_i)^3}
    \end{equation}
    where $\mu_i$ is the mean of the channel, $N$ is the total pixel count, and $p_{ij}$ denotes the intensity of the $j$-th pixel in channel $i$.
\end{enumerate} 
By concatenating these three moments for all three color channels, we generate a 9-dimensional feature vector.
\subsubsection{RGB Color Histograms} To capture the frequency distribution of pixel intensities across the color spectrum, we extracted grouped color histograms from the RGB channels. Unlike statistical moments, which summarize the distribution into single values, histograms preserve the density of pixel counts at specific intensity intervals. The feature extraction process involves three key steps:
\begin{enumerate}
\item Grouping: The standard 8-bit color depth provides 256 distinct intensity values per channel. To reduce feature dimensionality and mitigate the effect of minor noise, we quantized the intensity range $[0, 256]$ into $B$ bins. We experimentally evaluated bin sizes of $\{8, 16, 32, 64\}$ and determined that $B = 16$ yielded the highest classification accuracy. Consequently, the final model utilizes 16 bins, where each bin aggregates the pixel counts for a sub-range of intensity levels.
\item Histogram Calculation and Normalization: For each color channel $c \in \{R, G, B\}$, we computed the histogram $H_c$, where the value of the $k$-th bin represents the count of pixels falling within that intensity range. To ensure invariance to image resolution (image size), the histograms were normalized to represent probability distributions:$$P_c(k) = \frac{H_c(k)}{\sum_{j=1}^{B} H_c(j)}$$ This results in a feature vector of size $16$ per channel.
\item Integration of Global Statistics:
To reinforce the histogram data with global intensity trends, we explicitly calculated the Mean ($\mu$) and Standard Deviation ($\sigma$) for each channel and appended them to the vector. The final feature representation is a concatenation of the normalized histograms and statistical moments for all three channels:$$V = [H_R, H_G, H_B, \mu_R, \sigma_R, \mu_G, \sigma_G, \mu_B, \sigma_B]$$The resulting dimensionality of the feature vector is $(16 \times 3) + 6 = 54$ dimensions.
\end{enumerate}
\subsubsection{HSV Color Space Representations} The standard RGB model mixes the actual color with the brightness of the light. This is a problem in histopathology because a change in scanner lighting can change the RGB numbers, even if the biology remains the same. To fix this, we converted the images into the HSV (Hue, Saturation, Value) format. This separates the information into three clear parts: Hue tells us the actual color (distinguishing the purple Hematoxylin from the pink Eosin); Saturation measures how strong or deep the stain is; and Value measures the brightness, regardless of the color. 
\subsection{Classification Models}
To evaluate the discriminatory power of the extracted color features, we employed four distinct machine learning classifiers, each representing a different algorithmic strategy. We began with K-Nearest Neighbors (KNN) \cite{peterson2009k}, a simple instance-based algorithm that classifies a new image by identifying the most similar examples in the training set. This serves as a baseline to test whether images with similar color distributions naturally cluster together. We also utilized Support Vector Machines (SVM) \cite{hearst1998support}, which attempts to find the optimal boundary that separates the classes with the widest possible margin. This method is particularly useful for determining if the color features can be mathematically separated in a high-dimensional space.

To capture more complex, non-linear relationships in the data, we implemented two tree-based ensemble methods. Random Forest (RF) \cite{breiman2001random} constructs a multitude of decision trees during training and averages their outputs to improve accuracy and control overfitting. Finally, we employed Extreme Gradient Boosting (XGBoost) \cite{chen2015xgboost}, an advanced algorithm that builds trees sequentially, where each new tree specifically corrects the errors made by the previous ones. These ensemble methods are crucial for this study, as they help identify which specific color bins or statistical moments are the most important drivers of classification.
\begin{figure*}[ht]
    \centering
    \includegraphics[width=\textwidth]{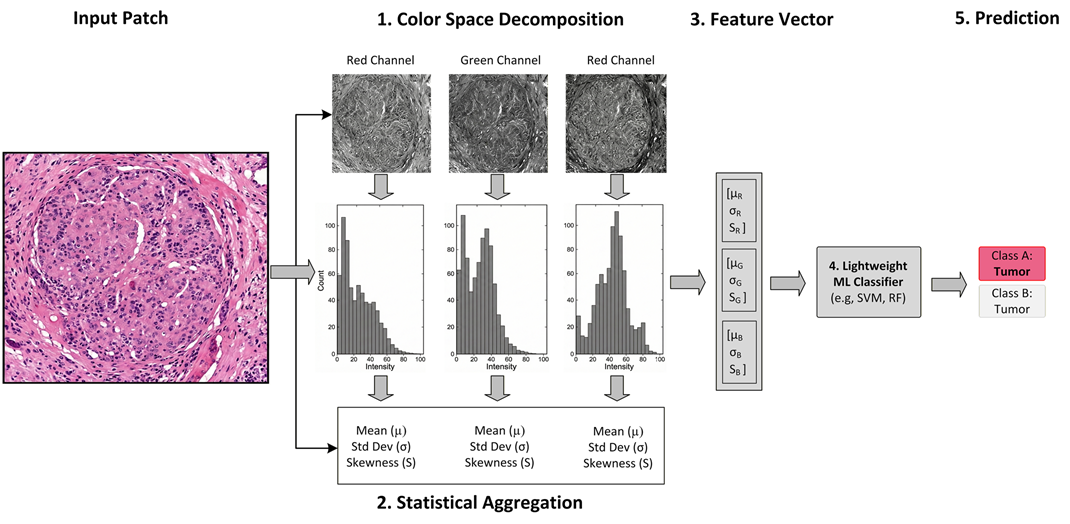} 
    \caption{Workflow for the Color Moments feature extraction and classification. An input H\&E histopathology patch is decomposed into its constituent color channels (Step 1). Spatial information is discarded, and the pixel intensity distribution for each channel is summarized using three statistical moments: Mean ($\mu$), Standard Deviation ($\sigma$), and Skewness ($S$) (Step 2). These statistics are concatenated across the three channels to form a compact, 9-dimensional feature vector (Step 3). This low-dimensional vector serves as the input for a lightweight machine learning classifier (e.g., SVM or RF) to determine the pathological class (Steps 4 and 5).}
    \label{fig:color_moment_workflow}
\end{figure*}

\section{Experimental Results}
Table \ref{tab:results} presents the complete performance benchmark across all 10 experimental setups. For each data set, we report the Balanced Accuracy achieved by the three feature extraction methods (Color Moments, RGB histograms, and HSV Histograms) across four distinct classifiers (KNN, SVM, Random Forest and XGBoost). The results demonstrate significant variability depending on tissue type, with Random Forest generally emerging as the most robust classifier across the majority of tasks. To further illustrate the practical potential of this procedure, Table \ref{tab:best_results_summary} summarizes the optimal configuration for each specific diagnostic task. This "best-of-each-case" approach simulates a clinical application that dynamically selects the most effective feature-classifier pair for a given tissue type.

\begin{table*}[h!]
\centering
\resizebox{\textwidth}{!}{%
\begin{tabular}{l l l | c c c | c c c | c c c | c c c}
\toprule
\multicolumn{3}{c|}{\textbf{Dataset \& Task Details}} & \multicolumn{3}{c|}{\textbf{KNN}} & \multicolumn{3}{c|}{\textbf{SVM}} & \multicolumn{3}{c|}{\textbf{Random Forest}} & \multicolumn{3}{c}{\textbf{XGBoost}} \\
\cmidrule(lr){1-3} \cmidrule(lr){4-6} \cmidrule(lr){7-9} \cmidrule(lr){10-12} \cmidrule(lr){13-15}
\textbf{Dataset} & \textbf{Classes} & \textbf{Task Category} & \textbf{Mom.} & \textbf{RGB} & \textbf{HSV} & \textbf{Mom.} & \textbf{RGB} & \textbf{HSV} & \textbf{Mom.} & \textbf{RGB} & \textbf{HSV} & \textbf{Mom.} & \textbf{RGB} & \textbf{HSV} \\
\midrule

PathMNIST & 2 & Diagnostic Class & 84\% & 82\% & 86\% & 76\% & 69\% & 84\% & 85\% & 85\% & \textbf{87\%} & 81\% & 81\% & 86\% \\ 
PathMNIST & 9 & Tissue Types & 68\% & 63\% & 69\% & 66\% & 57\% & 71\% & 70\% & 71\% & \textbf{74\%} & 70\% & 69\% & 72\% \\ \midrule

BreakHis 40$\times$ & 2 & Diagnostic Class & 63\% & 63\% & 53\% & 75\% & \textbf{77\%} & 56\% & 63\% & 60\% & 69\% & 63\% & 61\% & 63\% \\
BreakHis 40$\times$ & 4 & Pathological Subtypes &  26\% & 23\% & 29\% & 26\% & 20\% & 33\% & \textbf{38\%} & 35\% & 36\% & 27\% & 34\% & 26\% \\ \midrule

BreakHis 100$\times$ & 2 & Diagnostic Class & 71\% & 72\% & 71\% & 67\% & 71\% & 80\% & 73\% & 79\% & \textbf{89\%} & 71\% & 79\% & 82\%  \\
BreakHis 100$\times$ & 4 & Pathological Subtypes & 32\% & \textbf{34\%} & 33\% & 32\% & 30\% & 33\% & 32\% & 34\% & 33\% & 31\% & 30\% & 32\%  \\ \midrule

BreakHis 200$\times$ & 2 & Diagnostic Class & 34\% & 34\% & 33\% & 36\% & 35\% & 31\% & 30\% & 35\% & \textbf{39\%} & 37\% & 36\% & 32\% \\
BreakHis 200$\times$ & 4 & Pathological Subtypes & 32\% & 34\% & 33\% & 32\% & 30\% & 33\% & 32\% & \textbf{35\%} & 33\% & 31\% & 30\% & 32\% \\ \midrule

LungHist700 & 3 & Pathological Subtypes & 68\% & 62\% & 59\% & 34\% & 33\% & 68\% & 75\% & \textbf{79\%} & 71\% & 75\% & 75\% & 69\% \\
\midrule
IDC & 2 & Diagnostic Class &  79\% & 77\% & 74\% & \textbf{85\%} & \textbf{85\%} & 78\% & 79\% & 80\% & 72\% & 83\% & 81\% & 73\% \\

\midrule
\multicolumn{3}{c|}{\textbf{Overall Mean $\pm$ STD}} & 55.7$\pm$20 & 54.4$\pm$21 & 54.0$\pm$21 & 52.9$\pm$20 & 50.7$\pm$22 & 56.9$\pm$21 & 57.7$\pm$20 & 59.3$\pm$19 & \textbf{60.3$\pm$22} & 56.9$\pm$19 & 57.6$\pm$18 & 56.7$\pm$21 \\
\bottomrule
\end{tabular}%
}
\caption{Comprehensive experimental results reporting balanced accuracy across 10 experimental setups. We compare three feature extraction methods (Moments, RGB Hist, HSV Hist) across four classifiers. The final row represents the global average performance across all tasks.}
\label{tab:results}
\end{table*}

\begin{table}[h!]
\centering
\resizebox{\textwidth}{!}{
\begin{tabular}{l l l l c}
\toprule
\textbf{Dataset} & \textbf{Task Category} & \textbf{Best Classifier} & \textbf{Best Feature} & \textbf{Accuracy} \\
\midrule
PathMNIST & Diagnostic Class & Random Forest & HSV & 87\% \\
PathMNIST & Tissue Types & Random Forest & HSV & 74\% \\
\midrule
BreakHis 40$\times$ & Diagnostic Class & SVM & RGB Hist & 77\% \\
BreakHis 40$\times$ & Pathological Subtypes & Random Forest & Moments & 38\% \\
\midrule
BreakHis 100$\times$ & Diagnostic Class & Random Forest & HSV & 89\% \\
BreakHis 100$\times$ & Pathological Subtypes & KNN & RGB Hist & 34\% \\
\midrule
BreakHis 200$\times$ & Diagnostic Class & Random Forest & HSV & 39\% \\
BreakHis 200$\times$ & Pathological Subtypes & Random Forest & RGB Hist & 35\% \\
\midrule
LungHist700 & Pathological Subtypes & Random Forest & RGB Hist & 79\% \\
\midrule
IDC & Diagnostic Class & SVM & Mom. / RGB & 85\% \\
\bottomrule
\end{tabular}%
}
\caption{Optimized performance using task-specific model selection. This table reports the highest achieved balanced accuracy for each dataset by selecting the optimal combination of feature extraction method and classifier.}
\label{tab:best_results_summary}
\end{table}

\subsection{Diagnostic Capability in Binary Classification}
The binary classification experiments were designed to assess the baseline discriminative capacity of global color features in the complete absence of morphological information. Rather than constructing a sophisticated diagnostic pipeline, the objective was to determine whether pathological states (Normal vs. Abnormal or Benign vs. Malignant) exhibit sufficiently distinct color distributions to be separable using standard linear and non-linear classifiers. This analysis therefore isolates the contribution of pixel-level chromatic information to diagnostic performance. Across multiple datasets, the results demonstrate that global color statistics can serve as a strong proxy for malignancy in certain settings. In the PathMNIST binary task, balanced accuracies reached up to 87\%, while in the IDC dataset, classifiers achieved up to 85\% balanced accuracy using only color-based features (see Table \ref{tab:results}). These findings suggest that the presence of malignancy is often accompanied by systematic shifts in global color distributions, which are informative enough to be captured by classical classifiers such as Random Forests and Support Vector Machines, without reliance on deep learning based approaches or explicit shape and structural analysis.
\paragraph{The Effect of Magnification.}A critical finding of this study is that the diagnostic utility of color is highly dependent on magnification levels. Color features proved particularly effective for the 100$\times$ subset, where the Random Forest classifier achieved a peak balanced accuracy of 89\%. This indicates that, at this magnification, chromatic information alone contains sufficient discriminative signal to reliably separate benign from malignant tissue samples. In contrast, performance deteriorated markedly for the 200$\times$ subset, with balanced accuracies dropping to approximately 35\%, close to random chance. This pronounced performance gap highlights an important limitation of global color features: their diagnostic utility is not universal, but strongly dependent on imaging conditions such as magnification. While color statistics can be highly informative at certain scales, they become insufficient when chromatic differences are reduced by fine-grained structural details.

\subsection{Diagnostic Capability in Multi-class Classification}
The effectiveness of color-based features in multi-class settings varies significantly depending on the biological nature of the target classes. As summarized in Table \ref{tab:best_results_summary}, the framework transitions from high performance in broad tissue identification to lower accuracy in fine-grained subtyping.

The results indicate that the type of tissue being classified is the primary predictor of success for color-based models. In the PathMNIST dataset, which involves distinguishing between nine fundamentally different tissue types (e.g., Adipose, Muscle, Stroma), color-based representations proved highly effective. Because these tissues have distinct chemical compositions and react differently to H\&E staining, they produce highly separable color histograms. Using HSV features, the Random Forest classifier achieved a balanced accuracy of 74\%, far exceeding the 11\% random baseline. This demonstrates that global color statistics are highly reliable for identifying broad tissue phenotypes.

In contrast, when the task shifts to Pathological Subtyping (BreaKHis 4-class), the discriminative power of color is severely limited. These subtypes share similar biological origins and staining behaviors, differing primarily in architectural arrangement rather than pigment intensity. Consequently, accuracy for BreaKHis subtypes remained low, peaking at 38\%. The LungHist700 dataset represents a middle ground; it differentiates between benign tissue and two malignant subtypes (Adenocarcinoma and Squamous Cell Carcinoma). The model achieved a robust 79\% accuracy here, suggesting that while the two cancer subtypes share some color traits, their divergence from benign tissue and their unique inter-type staining intensities are sufficient for a color-based model to capture.

To address this variability, Table \ref{tab:best_results_summary} highlights that an adaptive application—selecting the optimal feature and classifier specifically for the tissue type and task at hand that can significantly improve performance. By tailoring the model to the biological context, the framework achieves an overall mean accuracy of 60.3\% $\pm$ 22\%. This suggests that while color features have limitations in fine-grained subtyping, they remain a powerful and efficient tool for broader tissue classification and diagnostic screening.

\section{Discussion}
This study examined the diagnostic value of global color features in histopathological images by intentionally excluding morphological information and spatial structure. By isolating chromatic cues, we aimed to quantify how much discriminative signal can be attributed to color alone, independent of cell shape, tissue architecture, or contextual relationships. The experimental results show that color-based representations can be surprisingly effective in certain classification settings. In binary tasks, such as distinguishing normal from abnormal tissue or benign from malignant samples, global color features consistently achieved balanced accuracies well above random chance and, in some cases, approached performance levels typically associated with more complex models. This suggests that malignancy is often accompanied by systematic changes in staining appearance that manifest as shifts in global color distributions. However, the effectiveness of color features was not uniform across datasets or imaging conditions. A particularly notable finding emerged from the BreaKHis dataset, where strong performance was observed at $100\times$ magnification, reaching 89\% with Random Forest, but deteriorated sharply at $200\times$ magnification. This behavior indicates that the discriminative power of global color statistics is sensitive to scale. At lower magnification, color variations tend to reflect broader tissue-level characteristics, whereas at higher magnification, diagnostically relevant information becomes increasingly localized and structural, limiting the usefulness of global color summaries.

The multi-class experiments further clarified the boundaries of color-based discrimination. High performance on PathMNIST can be attributed to the fact that its classes correspond to distinct tissue phenotypes with markedly different biological composition and staining behavior. In this context, color serves as an effective proxy for tissue identity. In contrast, the BreakHis subtype classification task, which involves separating closely related malignant subtypes, proved far more challenging. The low accuracy observed in this setting confirms that global color features are insufficient for fine-grained cancer subtype differentiation, as these distinctions rely primarily on architectural and morphological patterns rather than chromatic differences. The LungHist700 results demonstrates an intermediate position which reflects partial separability between benign tissue and distinct cancer types, but also highlighting the limitations of color alone for detailed characterization.

A critical outcome of this study is the ranking of methods based on the trade-off between computational efficiency and diagnostic accuracy. In terms of raw accuracy, the HSV Color Histogram paired with a Random Forest classifier is ranked as the top-performing configuration, as it achieved the highest overall mean accuracy of 60.3\% $\pm$ 22\% across all experimental tasks. Regarding memory usage and speed of computation, the Color Moments approach is ranked highest for efficiency because it reduces the input data to a compact 9-dimensional feature vector, whereas the histogram-based method requires 54 dimensions. This reduction in dimensionality significantly lowers the memory footprint, making it the most suitable choice for "TinyML" hardware. Overall, we rank the HSV-based Random Forest as the most robust choice for accuracy-critical tasks and the Color Moments-based SVM as the most efficient for resource-constrained environments where processing speed is the primary constraint.

These findings have important implications for the interpretation of machine learning models in computational pathology. While deep learning approaches integrate both color and morphological information, the strong performance achieved using color-only features suggests that chromatic cues can contribute to model predictions. This highlights the potential of using computationally efficient color statistics as a robust baseline for histopathological analysis. If simple chromatic features can achieve high diagnostic accuracy in specific tasks, they may serve as effective, low-cost screening tools for preliminary triage or quality control which may offer a transparent alternative for simpler diagnostic questions. Furthermore, as the reliability of color-based features is inherently linked to the consistency of the staining process, it is possible that future developments in standardized or improved histological staining methods could lead to even higher classification accuracies by increasing the color-based separation between different disease categories. In conclusion, this study demonstrates that while global color distribution is not a replacement for morphological analysis, it constitutes a powerful and independent diagnostic signal. For binary diagnostic decisions and distinct tissue phenotyping, raw color statistics offer a surprisingly strong proxy for malignancy. However, the inability of these features to distinguish between carcinoma subtypes confirms that the fine-grained characterization of cancer remains the exclusive domain of structural analysis. Future research in computational pathology should consider these chromatic baselines to distinctively quantify the contribution of color versus shape in automated diagnosis.

\bibliographystyle{unsrt}  
\bibliography{references}

\end{document}